\documentclass[10pt, oneside]{article}   	% use "amsart" instead of "article" for AMSLaTeX format
\pdfoutput=1
\usepackage{geometry}                		% See geometry.pdf to learn the layout options. There are lots.
\usepackage{graphicx}				% Use pdf, png, jpg, or eps with pdflatex; use eps in DVI mode						
\usepackage{amssymb}

\title{Analysis of the South Slavic Scripts by Run-Length Features of the Image Texture}
\author{Darko Brodi\'c \footnote{(Corresponding author) University of Belgrade, Technical Faculty in Bor, Vojske Jugoslavije 12, 19210 Bor, Serbia, dbrodic@tf.bor.ac.rs}, Zoran N. Milivojevi\'c \footnote{College of Applied Technical Sciences, Aleksandra Medvedeva 20, 18000 Ni\v s, Serbia, zoran.milivojevic@vtsnis.edu.rs}, Alessia Amelio \footnote{Institute for High Performance Computing and Networking, National Research Council of Italy, CNR-ICAR, Via P. Bucci 41C, 87036 Rende (CS), Italy, amelio@icar.cnr.it}}
\date{}							% Activate to display a given date or no date

\begin{document}
\maketitle
\begin{center}
\bf Abstract
\end{center}
The paper proposes an algorithm for the script recognition based on the texture characteristics. The image texture is achieved by coding each letter with the equivalent script type (number code) according to its position in the text line. Each code is transformed into equivalent gray level pixel creating an 1-D image. Then, the image texture is subjected to the run-length analysis. This analysis extracts the run-length features, which are classified to make a distinction between the scripts under consideration. In the experiment, a custom oriented database is subject to the proposed algorithm. The database consists of some text documents written in Cyrillic, Latin and Glagolitic scripts. Furthermore, it is divided into training and test parts. The results of the experiment show that 3 out of 5 run-length features can be used for effective differentiation between the analyzed South Slavic scripts.

{\it Index, Terms} - Classification, Coding, Image Analysis, Script, Texture.

\section{Introduction}

The Balkan region, which is populated by South Slavs, is very rich in cultural heritage elements of the medieval age. One of the most important cultural achievements represents the variety of used scripts. In the medieval age, South Slavs had spoken the old Church Slavonic language. It was written with the Glagolitic alphabet called round Glagolitic script, but later it was replaced by Cyrillic in the east region of Balkan, i.e. in Bulgaria and Macedonia. In Bosnia, the local version of Cyrillic alphabet was used, while in Croatia a variant of the Glagolitic alphabet called squared Glagolitic script was preserved. Accordingly, all books from medieval age were written by all aforementioned scripts. Currently, Serbian language is the only European standard language with complete synchronic digraphia, which uses both Cyrillic and Latin alphabets. 

The Serbian language has been studied in a natural speech analysis as a part of the South Slavic language group \cite{Sovi2014}. Similarly, the recognition of the South Slavic scripts can be used in document image analysis (DIA) and optical character recognition (OCR) \cite{Kast2002}. Recognition of different script characters in an OCR module is a difficult task \cite{Gho2010}. It is because the character recognition depends on different features like the structural properties, style and nature of writing, which generally differ from one script to another. It is especially true for digitalization and script recognition in the old books from the Balkan region, which are written by different scripts, i.e. Cyrillic, Glagolitic and Latin scripts.

The script recognition techniques have been classified as global or local. Global methods treat the document image as a group of big image blocks. Then, the image blocks are statistically analyzed \cite{Joshi2007}. The drawback of these methods corresponds to process the noise images, which can decay the recognition results \cite{Busch2005}. Local approaches segment document images into small blocks representing text pieces, i.e. connected components. Then, they are subjected to the statistical analysis like the run-length, co-occurrence, local binary patterns, etc \cite{Pal2002}. The efficiency of the local methods is unaffected by the noise in the document image. Unfortunately, they are more computer time intensive than global methods. 

This paper presents a method similar to local methods. It is based on coding each letter in the script by its position in the text line \cite{Busch2005}, \cite{Pal2002}, \cite{Brodic2013}. Accordingly, the method takes into account energy profile of each script sign in a certain text line and classifies it into four different groups \cite{Gho2010}. Hence, the efficiency of the method is linked with the basis of the printed text.  In this way, the number of initial variables is considerably reduced compared to the initial ones. The received codes are translated into the gray level pixels of the image. Then, the texture of the image is subjected to the run-length analysis. Extracted run-length features are the basis for distinguishing the scripts. This goal is accomplished by the classification tool GA-ICDA, an extension of GA-IC framework \cite{Amelio2014}, which segments the data into clusters representing different scripts. 

The remainder of the paper is organized as follows. Section 2 addresses all aspects concerning the proposed algorithm. It includes the text-line definition and script modeling. The result represents the coded text which corresponds to 1-D four gray level image. Furthermore, it is subjected to run-length analysis of the image texture. Then, the obtained results are analyzed and classified by the clustering algorithm. Section 3 describes the experiment and custom oriented database consisting in training and test document sets written in different scripts. Section 4 gives the results of the experiment and discusses them. Section 5 makes conclusions and points out further research direction.

\section{Algorithm}

\begin{figure}[!ht]
\begin{center}
\includegraphics[width=12cm, height=12cm, keepaspectratio]{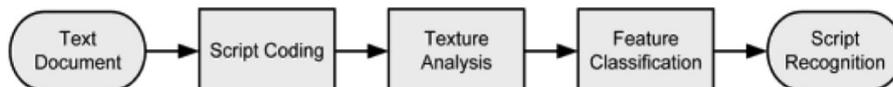}
\caption{The stages of the proposed algorithm.}
\label{figure1}
\end{center}
\end{figure}

The algorithm consists in the following three stages: (i) script coding, (ii) texture analysis, and (iii) classification. Script coding translates each letter into the gray-level pixel of the image in accordance to the position in the text line. Then, the texture analysis process derives a 1-D image to extract run-length texture features. These features are subject to the classification process in order to cluster classes representing documents written in different scripts. Fig. \ref{figure1} shows the stages of the proposed algorithm.

\subsection{Script Coding}
Document text can be segmented into text lines. Furthermore, each text line can be separated taking into account the energy of the script signs \cite{Joshi2007} into the following virtual lines \cite{Zram98}: (i) top-line, (ii) upper-line, (iii) base-line, and (iv) bottom-line. These lines divide the text line area into three vertical zones \cite{Zram98}: (i) upper zone, (ii) middle zone, and (iii) lower zone. 

\begin{figure}[!ht]
\begin{center}
\includegraphics[width=8cm, height=8cm, keepaspectratio]{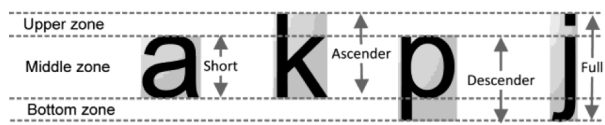}
\caption{Illustration of the vitual lines and vertical zones in the text line.}
\label{figure2}
\end{center}
\end{figure}

The letters can be classified depending on their position in vertical zones of the line, which reflects their energy profile. The short letters (S) occupy the middle zone only. The ascender letters (A) spread over the middle and upper zone. The descendent letters (D) enlarge into the middle and lower zone. The full letters (F) outspread over all vertical zones. Hence, all letters can be grouped into four different script types \cite{Brodic2013}. Fig. \ref{figure2} shows the script characteristics according to the letter baseline position.

\begin{figure}[!ht]
\begin{center}
\includegraphics[width=7cm, height=7cm, keepaspectratio]{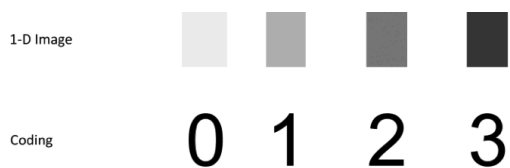}
\caption{Script type equivalent number codes and corresponding gray levels.}
\label{figure3}
\end{center}
\end{figure}

Each of the script types can be exchanged with different number codes: 0, 1, 2, 3. Accordingly, there exist only four script types, which lead to four number codes. Furthermore, to create a texture these numbers are transformed into different levels of gray. Fig. \ref{figure3} shows the equivalence between script type number codes and gray levels.

In this way, each text is transformed into the set of number codes (0, 1, 2, 3) and later into the pixels of only four gray levels. Obtained image is a 1-D image $I_m$ representing a texture, which can be subjected to the texture analysis. 

\subsection{Texture Analysis}
Texture is a measure of the intensity variation in the image surface. Hence, it is used to extract information from the images that can quantify the properties like smoothness, coarseness, and regularity \cite{Srin2008}. In this way, the texture is characterized by calculating statistical measures obtained from the grayscale intensities in the image. Aforementioned statistical measures can be used for classification and segmentation of the image \cite{Jain95}.  

Run-length statistical analysis is used to evaluate and quantify the coarseness of a texture \cite{Xu2004}. A run can be defined as a set of consecutive pixels, which are characterized by the same gray level intensity in a specific direction. Typically, coarse textures contain long runs with different gray level intensities, while fine textures include short runs with similar gray level intensities. 

Let's suppose that we have an image $I_m$ featuring X rows, Y columns and M levels of gray intensity. The starting point is the extraction of the run-length matrix p(i, j). It is defined as the number of runs with pixels of gray level i = 1,..., M and run length j = 1,..., N, where M is the number of gray levels, while N is the number representing the maximum run length. In our case, each element of the run-length matrix p(i, j) represents the gray level run-length of the 1-D matrix $I_m$ that gives the total number of occurrences of gray-level runs of length j and of intensity value i. A set of consecutive pixels with identical or similar intensity values constitutes a gray level run. Furthermore, various texture matrices and vectors representing semi-features can be extracted from this run-length matrix p(i, j) \cite{Tang98}: (i) gray level run-length pixel number matrix $p_p$, (ii) gray level run-number vector $p_g$, (iii) run-length run-number vector $p_r$, and (iv) gray level run-length-one vector $p_o$.

Gray level run-length pixel number matrix $p_p$ is defined as:

\begin{equation}
p_p(i,j)=p(i,j) \cdot j
\end{equation}

Gray level run-number vector $p_g$ is calculated as:

\begin{equation}
p_g(i,j)=\sum_{j=1}^N p(i,j)
\end{equation}

Run-length run-number vector $p_r$ is given as:

\begin{equation}
p_r(i,j)=\sum_{i=1}^M p(i,j)
\end{equation}

Gray level run-length-one vector $p_o$ is:

\begin{equation}
p_o(i,j)=p(i,1)
\end{equation}

Using aforementioned matrix and vectors, the following run-length features were originally proposed in \cite{Galloway75}: (i) Short run emphasis (SRE), (ii) Long run emphasis (LRE), (iii) Gray-level nonuniformity (GLN), (iv) Run length nonuniformity (RLN), and (v) Run percentage (RP).

SRE measures the distribution of short runs. It is highly dependent on the occurrence of short runs and is expected to be large for fine textures. It is given as:

\begin{equation}
SRE=\frac{1}{n_r}\sum_{i=1}^M \sum_{j=1}^N \frac{p(i,j)}{j^2} = \frac{1}{n_r}\sum_{j=1}^N \frac{p_r(j)}{j^2}
\end{equation}

LRE measures the distribution of long runs. It is highly dependent on the occurrence of long runs and is expected to be large for coarse structural textures. It is calculated as:

\begin{equation}
LRE=\frac{1}{n_r}\sum_{i=1}^M \sum_{j=1}^N {p(i,j)}\cdot {j^2} = \frac{1}{n_r}\sum_{j=1}^N {p_r(j)}\cdot {j^2}
\end{equation}

GLN measures the similarity of gray level values throughout the image. Its value is expected to be small if the gray level values are alike throughout the image. It is computed as:

\begin{equation}
GLN=\frac{1}{n_r}\sum_{i=1}^M (\sum_{j=1}^N {p(i,j)})^2= \frac{1}{n_r}\sum_{i=1}^M {p_g(i)}^2
\end{equation}

RLN measures the similarity of the length of runs throughout the image. It is expected to receive smaller values if the run lengths are alike throughout the image. It is defined as:

\begin{equation}
RLN=\frac{1}{n_r}\sum_{j=1}^N (\sum_{i=1}^M {p(i,j)})^2= \frac{1}{n_r}\sum_{j=1}^N {p_r(j)}^2
\end{equation}

RP measures the homogeneity and the distribution of runs of an image in a specific direction. It receives the largest values when the length of runs is 1 for all gray levels in a specific direction. It is given as:

\begin{equation}
RP = \frac{n_r}{n_p}
\end{equation}

In eqs. (5)-(9) $n_r$ represents the total number of runs, while $n_p$ is the number of pixels in the image $I_m$.

\subsection{Classification}

Obtained features can be analyzed in order to extract only run-length features that can distinguish different scripts. For this purpose, we adopt an extension of the GA-IC tool in \cite{Amelio2014}, called GA-ICDA (Genetic Algorithms Image Clustering for Document Analysis). GA-IC is used for clustering images from a database. It creates a weighted graph. The nodes of the graph correspond to images. An edge of the graph connecting two images is established if the images are similar. Each edge has its weight, which expresses the level of similarity extracted from the feature vectors of each image. The graph of images is then clustered by applying a genetic algorithm that divides the graph into groups of nodes. GA-ICDA introduces three new aspects from GA-IC. First of all, each graph node is a document represented as a run-length feature vector. Secondly, an edge connects two nodes if related documents are similar and also the node distance is less than a threshold T, established a node ordering. Finally, a refinement procedure at the end of the genetic algorithm merges pairs of clusters with minimum distance to each other, until a fixed number is reached. 

\section{Experiment}

The goal of the experiment is to prove the effectiveness of the proposed algorithm for the script recognition. Serbian or Croatian languages can be written by three different scripts: Cyrillic, Latin and Glagolitic. Hence, it is suitable for the experiment. For the purpose of the experiment, a custom oriented database of text documents written in Cyrillic, Latin and Glagolitic scripts is created. The database includes two parts: training and test set of documents. Training set consists in a total of 100 documents, while test set consists in a total of 15 documents. Each set incorporates Cyrillic, Latin and Glagolitic scripts in a similar number of documents. Both sets are subjected to the run-length statistical analysis. The obtained results are given and discussed below.

\section{Results and Discussion}

First, the training set is subjected to the run-length statistical analysis. The results are given in Tab. I for the Cyrillic, Latin and Glagolitic scripts.   

\begin{figure}[!ht]
\begin{center}
\includegraphics[width=9cm, height=9cm, keepaspectratio]{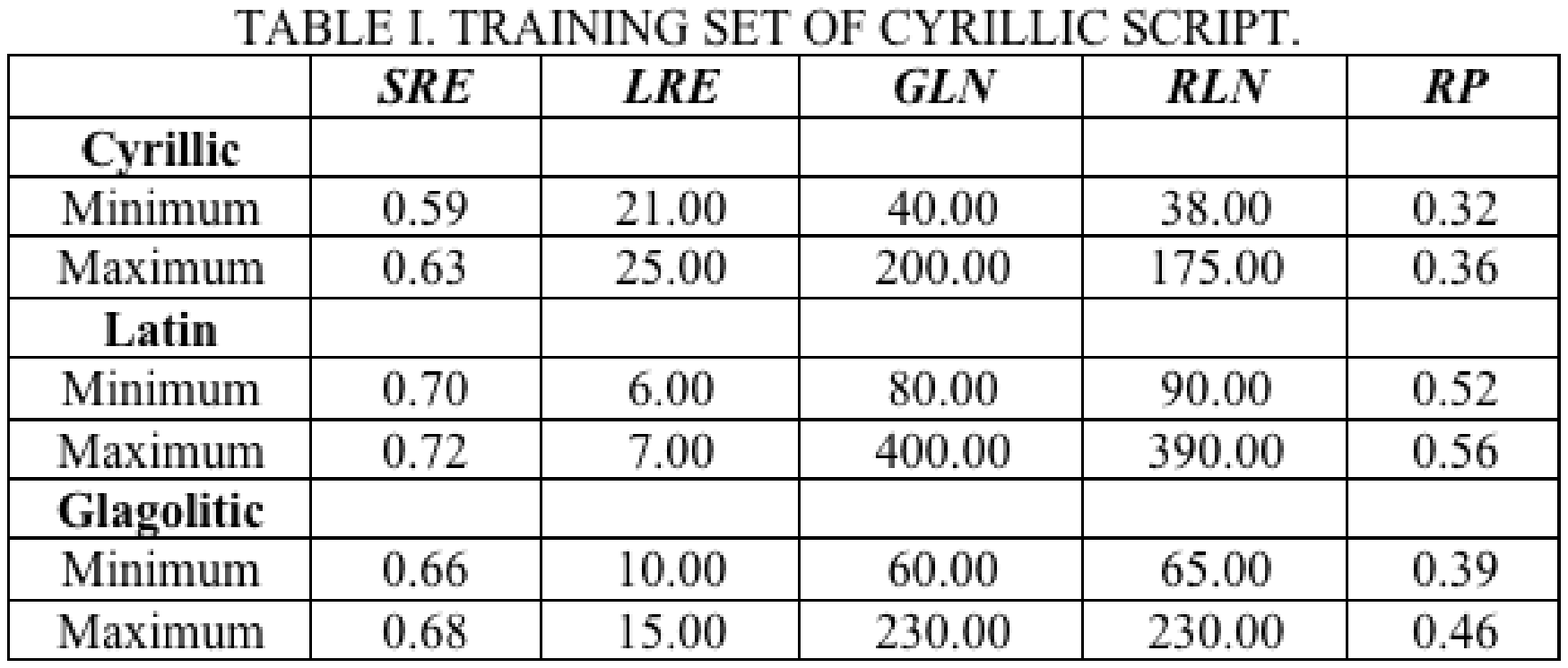}
\end{center}
\end{figure}

Then, the test set is subjected to the run-length statistical analysis. The results are given in Tab. II for the Cyrillic, Latin and Glagolitic scripts.

\begin{figure}[!ht]
\begin{center}
\includegraphics[width=9cm, height=9cm, keepaspectratio]{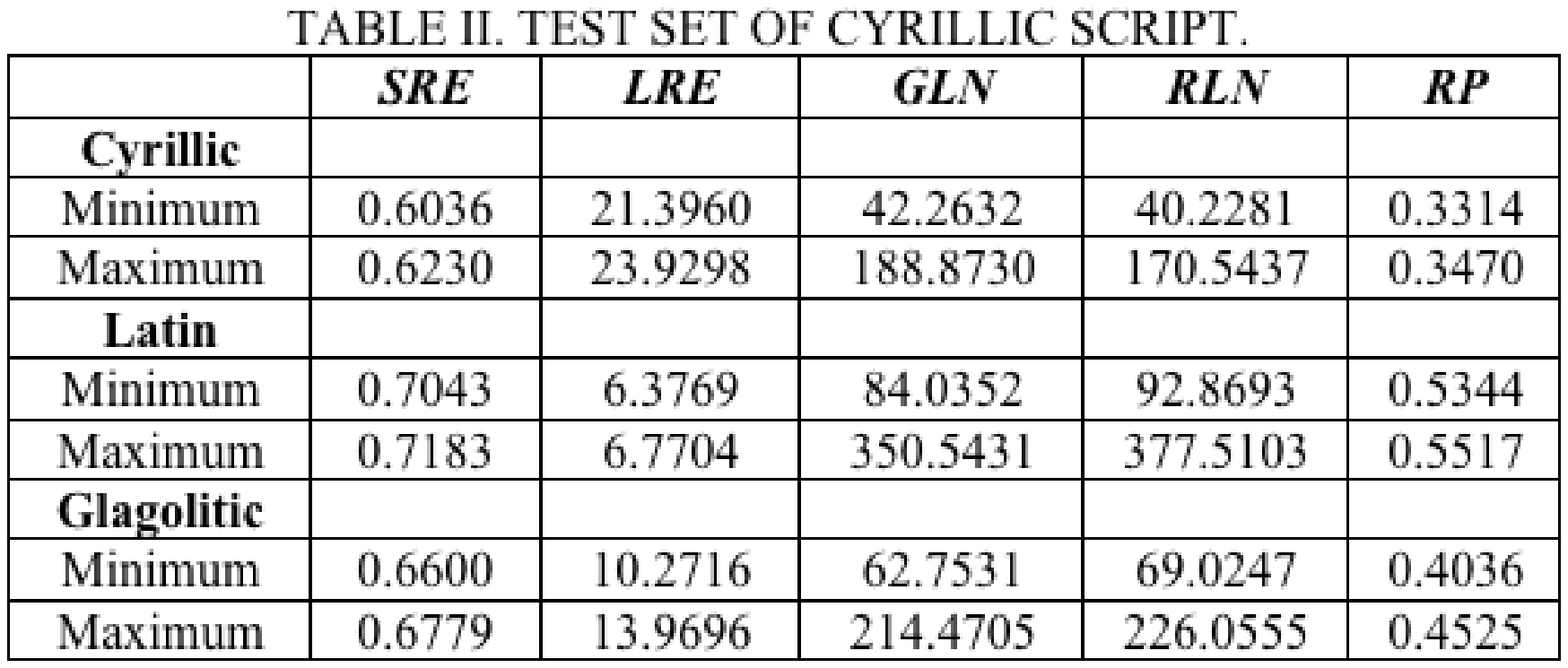}
\end{center}
\end{figure}

\begin{figure}[!ht]
\begin{center}
\includegraphics[width=8cm, height=8cm, keepaspectratio]{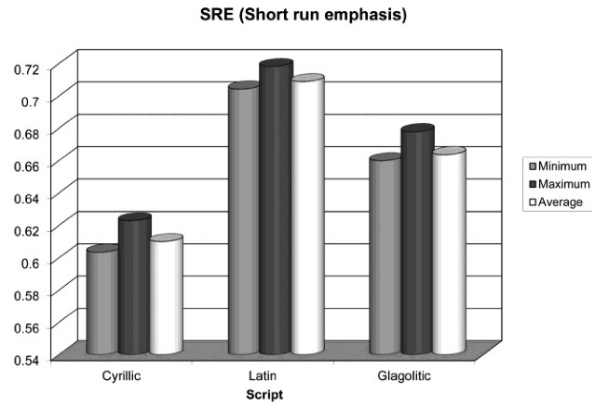}
\caption{SRE for test set of the scripts.}
\label{figure4}
\end{center}
\end{figure}

The above results represent the extracted five run-length features for each script. Currently, the classification problem is mandatory. First, a comparison of the training and test set results is important. If the test set results are a subset of the training set results, then the results are valid. Furthermore, we have to explore the run-length features that can be separated according to the given script. From Tabs. I-II, the values of GLN and RLN features are mutually overlapped between the scripts. However, SRE, LRE and RP characterize each script differently. Fig. \ref{figure4} shows an SRE test set of the scripts.

\begin{figure}[!ht]
\begin{center}
\includegraphics[width=8cm, height=8cm, keepaspectratio]{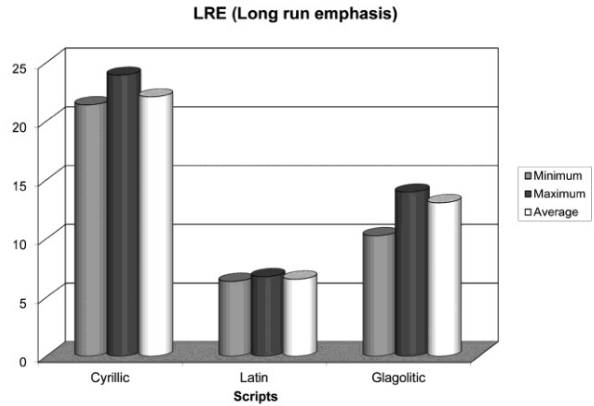}
\caption{LRE for test set of the scripts.}
\label{figure5}
\end{center}
\end{figure}

The SRE range given from minimum to maximum obtained values for each script is clearly distinct between each script. Hence, SRE is suitable for differentiation between the scripts. Fig. \ref{figure5} shows a LRE test set of the scripts.

\begin{figure}[!ht]
\begin{center}
\includegraphics[width=8cm, height=8cm, keepaspectratio]{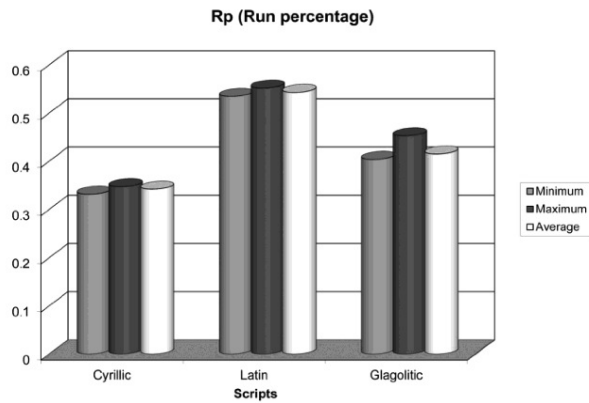}
\caption{RP for test set of the scripts.}
\label{figure6}
\end{center}
\end{figure}

The LRE range for each script can distinguish between each script. Accordingly, LRE is also suitable for distinction between scripts. Fig. \ref{figure6} shows an RP test set of the scripts.

The RP range for each script is quite distinct from each others. Hence, the combination of the SRE, LRE and RP can be used to freely distinct the characteristics of different scripts like South Slavic scripts: Cyrillic, Latin and Glagolitic. It represents a much easier solution than those given in \cite{Brodic2014}. GA-ICDA is used to classify the texture features and cluster data. 

Our goal is to classify a model and correctly predict the classes that represent documents written by different scripts. Hence, precision, recall and f-measure are preferred metrics to evaluate the proposed algorithm. Precision is the fraction of retrieved and relevant documents with respect to all retrieved documents. Recall is the fraction of relevant documents which are retrieved with respect to all relevant documents. F-Measure is the harmonic mean of precision and recall. Classification by GA-ICDA on the test set established 3 communities with the correct prediction of each document in the right class as given in Table III.

Hence, precision, recall and f-measure receive a value of 1. In this way, the proposed algorithm correctly predicts the classification of each document from a database in the adequate class representing Cyrillic, Latin or Glagolitic scripts (3 classes). Comparisons with other two well-known classifiers, Hierarchical Clustering and Expectation-Maximization, on the same run-length coded test set revealed the superiority of our approach. Each algorithm has been executed 100 times and the average values together
with standard deviation (in parenthesis) are reported.

\begin{figure}[!ht]
\begin{center}
\includegraphics[width=10cm, height=10cm, keepaspectratio]{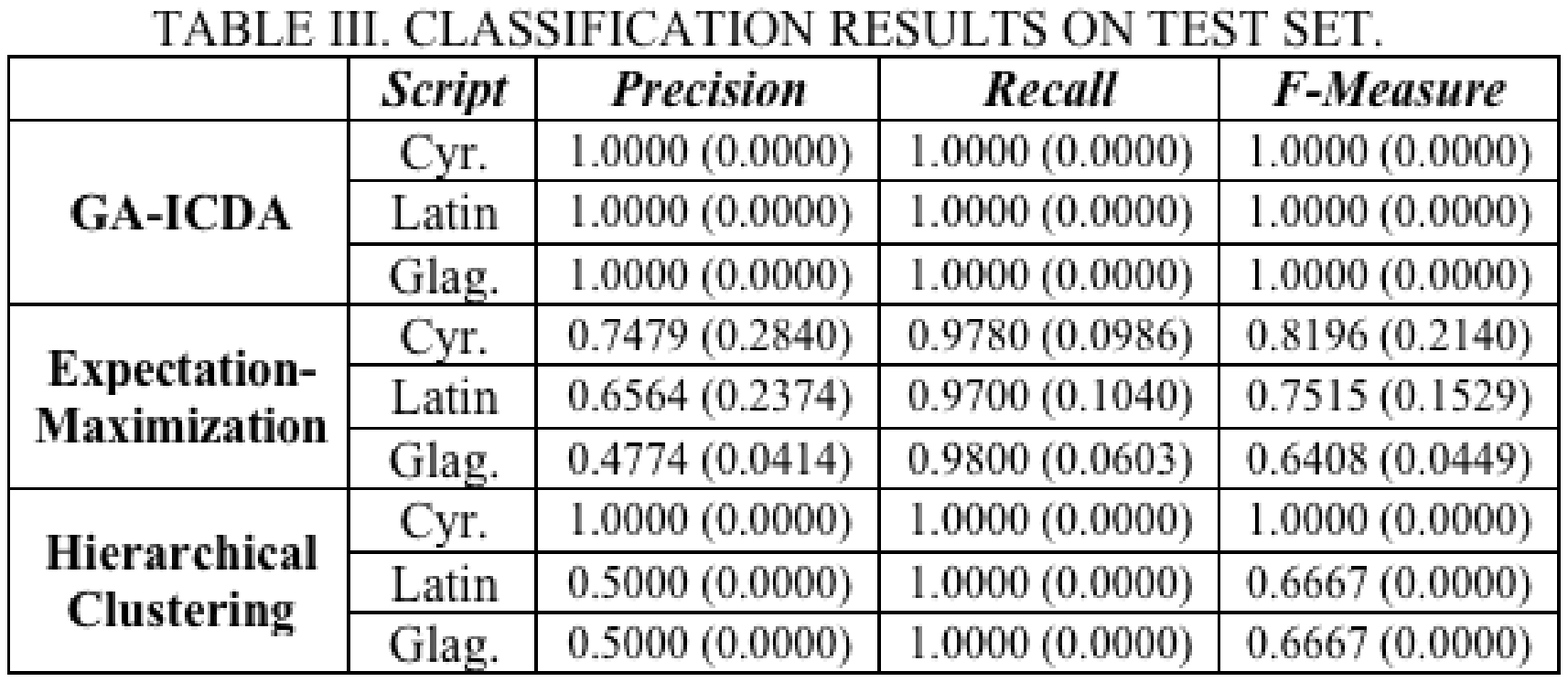}
\end{center}
\end{figure}

\section{Conclusions}
The manuscript proposed a methodology for the script recognition in the South Slavic documents. It uses the run-length statistical analysis of the document based on the status of each script element in the text line. Due to the difference in the script characteristics, the results of the statistical analysis show significant dissimilarity. It represents a starting point for feature classification. The proposed method is tested on documents written in Cyrillic, Latin and Glagolitic scripts. The experiments show encouraging results. Further research direction will be toward the statistical analysis of other scripts. \cite{Brodic2014}

\end{document}